\documentclass[12pt,twoside]{article}
\usepackage{latexsym,amssymb}
\usepackage{epsfig} %
\newtheorem{Theorem}{Theorem}

\def\baq#1\eaq{\begin{align}#1\end{align}}
\def\baqn#1\eaqn{\begin{align*}#1\end{align*}}
\def\beq#1\eeq{\begin{equation}#1\end{equation}}
\def\beqn#1\eeqn{\begin{displaymath}#1\end{displaymath}}
\def\bqa#1\eqa{\begin{eqnarray}#1\end{eqnarray}}
\def\bqan#1\eqan{\begin{eqnarray*}#1\end{eqnarray*}}

\newcommand{\calL}{\mathcal L}
\newcommand{\calT}{\mathcal T}
\newcommand{\calX}{\mathcal X}
\newcommand{\NNN}{\mathbb N}
\newcommand{\RRR}{\mathbb R}
\newcommand{\Expect}{{\mathbf E}}
\newcommand{\eps}{\varepsilon}
\newcommand{\leqt}{_{1:t}}
\newcommand{\ltt}{_{<t}}
\newcommand{\leqT}{_{1:T}}
\newcommand{\und}{\mbox{ and }}
\newcommand{\toinfty}[1]{\stackrel{#1\to\infty}{\longrightarrow}}
\newcommand{\tfrac}[2]{{\textstyle\frac{#1}{#2}}}
\newcommand{\wwwtilde}{$^{_{_\sim}}\!$}
\newcommand{\paradot}[1]{\vspace{1.3ex plus 0.5ex minus 0.5ex}\noindent{\bf{#1.}}}
\newcommand{\paranodot}[1]{\vspace{1.3ex plus 0.5ex minus 0.5ex}\noindent{\bf{#1}}}
\newcommand{\M}{{\cal M}}
\newcommand{\xiai}{{\xi}}
\newcommand{\text}[1]{\mbox{\scriptsize{#1}}}
\newcommand{\sfrac}[2]{{#1}/{#2}}
\newcommand{\xwidehat}{}
\newcommand{\FOE}{{\xwidehat{\mbox{\textit{F\kern-0.12emo\kern-0.08emE}}}}}
\newcommand{\FPL}{{\xwidehat{\mbox{\textit{F\kern-0.08emP\kern-0.08emL}}}}}
\newcommand{\foe}{^{\xwidehat{\mbox{\scriptsize\textit{F\kern-0.13emo\kern-0.13emE}}}}}
\newcommand{\fpl}{^{\xwidehat{\mbox{\scriptsize\textit{F\kern-0.15emP\kern-0.15emL}}}}}
\newcommand{\Loss}{\ell}
\newcommand{\loss}{\ell}
\newlength\algowidth
\newlength\graphwidth
\newcommand{\ii}{\hspace*{0.5em}}
\newcommand{\AIXI}{AI$\xi$}

\begin{document}

\title{\vspace{-3ex}\normalsize\sc Technical Report \hfill IDSIA-18-05
\vskip 2mm\bf\Large\hrule height5pt \vskip 6mm
Universal Learning of Repeated Matrix Games
\vskip 6mm \hrule height2pt}

\author{Jan Poland \\
\normalsize Grad.\ School of Inf.\ Sci.\ and Tech.\\
\normalsize Hokkaido University, Japan \\
\normalsize  \texttt{jan@ist.hokudai.ac.jp}\\
\normalsize  \texttt{www-alg.ist.hokudai.ac.jp/\wwwtilde jan}
\and
Marcus Hutter\\
\normalsize IDSIA, Galleria 2\\
\normalsize CH-6928 Manno (TI), Switzerland\\
\normalsize \texttt{marcus@idsia.ch}\\
\normalsize \texttt{www.idsia.ch/\wwwtilde marcus}
}

\date{16. August 2005}

\maketitle

\begin{abstract}
We study and compare the learning dynamics of two universal
learning algorithms, one based on Bayesian learning and the
other on prediction with expert advice. Both approaches have
strong asymptotic performance guarantees. When confronted with
the task of finding good long-term strategies in repeated
$2\times 2$ matrix games, they behave quite differently.
\end{abstract}

%%%%%%%%%%%%%%%%%%%%%%%%%%%%%%%%%%%%%%%%%%%%%%%%%%%%%%%%%%%%%%%
\section{Introduction}\label{secInt}
%%%%%%%%%%%%%%%%%%%%%%%%%%%%%%%%%%%%%%%%%%%%%%%%%%%%%%%%%%%%%%%

%------------------------------%
%\paradot{Problem specific versus universal learning}
%------------------------------%
Today, Data Mining and Machine Learning is typically treated in a
\emph{problem-specific} way: People propose algorithms to
solve a particular problem (such as learning to classify
points in a vector space), they prove properties and
performance guarantees of their algorithms (e.g.\ for Support
Vector Machines), and they evaluate the algorithms on toy or
real data, with the (potential) aim to use them afterwards in
real-world applications. In contrast, it seems that
\emph{universal learning}, i.e.\ a \emph{single} algorithm
which is applied for \emph{all} (or at least ``many")
problems, is neither feasible in terms of computational costs
nor competitive in (practical) performance. Nevertheless,
understanding universal learning is important: On the one
hand, its practical success would lead a way to Artificial
Intelligence. On the other hand, \emph{principles} and ideas
from universal learning can be of immediate use, and of course
Machine learning research aims at exploring and establishing
more and more general concepts and algorithms.

%------------------------------%
%\paradot{Universal learners}
%------------------------------%
Because of its practical restrictions, most of the understanding
of universal learning so far is theoretical. Some approaches which
have been suggested in the past are (adaptive) Levin search
\cite{Levin:73search,Schmidhuber:96}%
, Optimal Ordered Problem
Solver
\cite{Schmidhuber:02bias,Schmidhuber:04oops}
and Reinforcement Learning with split trees
\cite{Ring:94,McCallum:95}
among others. For a thorough discussion see e.g.\
\cite{Hutter:04uaibook}. In this paper, we concentrate on two
approaches with very strong theoretical guarantees in the
limit: the \AIXI\ agent based on Bayesian learning
\cite{Hutter:02selfopt} and
$\FOE$ based on Prediction with expert advice
\cite{Poland:05dule}.

%------------------------------%
%\paradot{Sequential decision theory}
%------------------------------%
Both models work in the setup of a \emph{sequential decision
problem}: An \emph{agent} interacts with an
\emph{environment} in discrete time $t$. At each time step, the
agent does some \emph{action} and receives a
\emph{feedback} from the environment. The feedback consists of
a \emph{loss} (or \emph{reward}) plus maybe more information.
(It is usually just a matter of convenience if losses or
rewards are considered, as one can be transformed into the
other by reverting the sign. Accordingly, in this paper we
switch between both, always preferring the more convenient
one.) In addition to this \emph{instantaneous} loss (or
reward), we will also consider the \emph{cumulative} loss
which is the sum of the instantaneous losses from $t=1$ up to
the current time step, and the \emph{average per round} loss
which is the cumulative loss divided by the total number of
time steps so far.

%------------------------------%
%\paradot{Passive versus reactive environments}
%------------------------------%
Most learning theory known so far concentrates on
\emph{passive} problems, where our actions have an influence
on the instantaneous loss, but \emph{not on the future
behavior of the environment}. All regression, classification,
(standard) time-series prediction tasks, common Bayesian
learning and prediction with expert advice, and many others
fall in this category. In contrast, here we deal with
\emph{active} problems. The environment may be
\emph{reactive}, i.e.\ react to our actions, which is the standard
situation considered in Reinforcement Learning. These cases
are harder in theory, and it is often impossible to obtain
relevant performance bounds in general.

%------------------------------%
%\paradot{Model classes}
%------------------------------%
Both approaches we consider and compare are based on finite or
countably infinite \emph{base classes}. In the Bayesian
decision approach, the base class consists of
\emph{hypotheses} or \emph{models} for the environment. A
model is a complete description of the (possibly
probabilistic) behavior of the environment. In order to prove
guarantees, it is usually assumed that the true environment is
contained in the model class. Experts algorithms in contrast
work with a class of decision-makers or \emph{experts}.
Performance guarantees are proven without any assumptions
\emph{in the worst case}, but only relative to the best
expert in the class. In both approaches, the model class is
endowed with a \emph{prior}. If the model class is finite and
contains $n$ elements, it is common to choose the uniform
prior $\frac{1}{n}$. For \emph{universal} learning it turns
out that universal base classes for both approaches can be
constructed from the set of all programs on some fixed
universal (prefix) Turing machine. Then each program naturally
corresponds to an element in the base class, and a prior
weight is defined by $w(\mbox{\textit{program}}) =
2^{-\mbox{\scriptsize\textit{length}(\textit{program})}}$
(provided that the input tape of the Turing machine is
binary). The prior is a (sub-)probability distribution on the
class, i.e.\ $\sum w\leq 1$.

%------------------------------%
\paradot{Contents}
%------------------------------%
The aim of this paper is to better understand the \emph{actual
learning dynamics} and properties of the two universal
approaches, which are both ``universally optimal" in a sense
specified later. Clearly, the universal base class is
computationally very expensive or infeasible to use. So we
will restrict on simpler base classes which are ``universal"
in a much weaker sense: we will employ complete \emph{Markov}
base classes where each element sees only the previous time
step. Although these classes are not truly universal, they are
general enough (and not tailored towards our applications),
such that we expect the outcome to be a good indication for
the dynamics of true universal learning. The problems we study
in this paper are \emph{$2\times 2$ matrix games}. (Due to lack of space,
we will not go into the deep literature on
learning equilibria in matrix games, as our primary interest
the universal learning dynamics.)
Matrix games are simple enough such that a ``universal"
algorithm with our restricted base class can learn something,
yet they provide interesting and nontrivial cases for reactive
environments, where really active learning is necessary.
Moreover, in this way we can set up a direct competition
between the two universal learners.
The paper is structured as follows: In the next two sections,
we present both universal learning approaches together with
their theoretical guarantees. Section \ref{secSim} contains
the simulations, followed by a discussion in Section
\ref{secDC}.

%%%%%%%%%%%%%%%%%%%%%%%%%%%%%%%%%%%%%%%%%%%%%%%%%%%%%%%%%%%%%%%
\section{\boldmath Bayesian Sequential Decisions (AI$\xi$)}\label{secAIXI}
%%%%%%%%%%%%%%%%%%%%%%%%%%%%%%%%%%%%%%%%%%%%%%%%%%%%%%%%%%%%%%%

%------------------------------%
\paradot{Passive problems}
%------------------------------%
Every inductive inference problem can be brought into the
following form: Given a string
$x_{<t}=x_{1:t-1}:=x_1x_2...x_{t-1}$, guess its
continuation $x_t$. Here and in the following we assume that
the symbols $x_t$ are in a finite alphabet $\calX$, for
concreteness the reader may think of $\calX=\{0,1\}$. If
strings are sampled from a probability distribution
$\mu:\calX^*\to [0,1]$, then predicting according to
$\mu(x_t|x\ltt)$, the probability conditioned on the history,
is optimal.  If $\mu$ is unknown, predictions may be based on
an approximation of $\mu$. This is what happens in
\emph{Bayesian sequential prediction}: Let the \emph{model
class} ${\M}\!:=\!\{\mu_1,\mu_2,...\}$ be a finite or
countable set of distributions on strings
$\mu_i(x\leqt|y\leqt)$ which are additionally conditionalized
to the past actions $y\ltt$. The actions are necessary for
dealing with sequential decision problems as introduced above.
We agree on the convention that the learner issues action
$y_t$ \emph{before} seeing $x_t$. Let $\{w_1,w_2,\ldots\}$ be a prior on
$\M$ satisfying $\sum w_i\leq 1$. Then the \emph{Bayes mixture} is the weighted
average
\beqn
  \xi(x_{1:t}|y\leqt) \!:=\!\!
  \sum\nolimits_iw_i\cdot\mu_i(x_{1:t}|y\leqt).
\eeqn
One can show that the $\xi$-predictions rapidly converge to
the $\mu$-predictions almost surely, if we assume that $\M$
contains the true distribution: $\mu\!\in\!\M$. This is not a
serious constraint if we include {\it all} computable
probability distributions in $\M$. This universal model class
corresponds to all programs on a fixed universal Turing
machine (cf. the introduction and
\cite{Solomonoff:64,Hutter:04uaibook}).

In a passive prediction problem, the behavior of the environments
$\mu_i$ do not depend on our actions $y_{1:t}$. Here we may
interpret our action $y_t$ as the prediction of $x_t$. Assume that
$\ell:(y_t,x_t)\mapsto[0,1]$ is a function defining our
instantaneous loss. Then \emph{the average per round regret of
$\xi$ tends to 0 at least at rate $t^{-\sfrac{1}{2}}$}, precisely
\beq
\label{eq:lb}
\tfrac{1}{t}L^\xi\leqt\leq\tfrac{1}{t}L^\mu\leqt+2\sqrt{\ln
w_\mu^{-1}/t}.
\eeq
Here, $L^\xi\leqt$ is the cumulative $\mu$-expected loss of
the $\xi$-predictions. The $\xi$-prediction (and likewise the
$\mu$-prediction) is chosen \emph{Bayes optimal}
for the given loss function:
$y_t^\xi=\arg\min_{y_t}\sum_{x_t}\ell(y_t,x_t)\xi(x\leqt|y\leqt)$.
The difference $L^\xi\leqt-L^\mu\leqt$ is termed
\emph{regret}.

\paradot{Active problems}
If the environment is \emph{reactive}, i.e.\ depends on our
action, then it is easy to construct examples where the greedy
Bayes optimal loss minimization is not optimal. Instead, the
\emph{far-sighted \AIXI-agent} chooses the action
\beq\label{ydotxi}
  y_t^{\xi,d}=
  \arg\min_{y_t}\!\sum_{x_t}...
  \min_{y_{t+d}}\!\sum_{x_{t+d}}
  \ell_{t:t+d}\xiai(x_{1:t+d}|y_{1:t+d}).
\eeq%
where
$\ell_{t:t+d}=\ell(y_{t:t+d},x_{t:t+d})=\sum_{s=t}^{t+d}\ell(y_s,x_s)$
and $d$ is the depth of the \emph{expectimin-tree} the agent
computes by means of (\ref{ydotxi}). We refer to $t+d$ as the
(current) \emph{horizon}. If we knew the final time
$T$ in advance and had enough computational resources, we could
choose $d=T-t$ according to the \emph{fixed horizon} $T$.
Taking $d$ fixed and small (e.g.
$d=8$) is computationally feasible, this is the \emph{moving
horizon} variant. However, this can cause consistency
problems: A sequence of actions which is started some time
step $t$ may not seem favorable any more in the next time step
$t+1$ (since the horizon shifts), and thus is disrupted. We
therefore also use an \emph{almost consistent horizon variant}
which takes $d=8$ in the first step, then $d=7$, and so on
down to $d=2$, after which we start again with $d=8$.
(Actually, we do not go down to $d=1$, since then the agent
would be greedy, which can for instance disrupt consecutive
runs of cooperation in the Prisoner's dilemma, see below.) A
theoretically very appealing alternative is to consider the
future \emph{discounted} loss and infinite depth, which is a
solution of the Bellman equations. This can be found in
\cite{Hutter:04uaibook}, together with more discussion and the
proof of the following \emph{optimality theorem} for \AIXI.

\begin{Theorem}[\boldmath Performance of AI$\xi$]\label{thSelfOpt}
If there exists a self-optimizing policy $\rho$ in the sense that
its expected average loss $\tfrac{1}{T} L_{1:T}^\rho$ converges
for $T\to\infty$ to the optimal average $\tfrac{1}{T} L_{1:T}^\mu$
for all environments $\mu\in\M$, then this also holds for the
universal policy $\xi$, i.e.\
\beqn
 \mbox{If }
  \exists\rho\forall\mu:
 \tfrac{1}{T} L_{1:T}^\rho \;\toinfty{T}\; \tfrac{1}{T} L_{1:T}^\mu
 \quad\Rightarrow\quad
 \tfrac{1}{T} L_{1:T}^\xi \;\toinfty{T}\; \tfrac{1}{T} L_{1:T}^\mu
\eeqn
\end{Theorem}

\begin{figure}[t!]
\begin{center}
\begin{minipage}{0.7\textwidth}
\begin{center}
\fbox{
\begin{minipage}{\textwidth}
\textbf{function}
$\ell=A(s_0,x\ltt,y\ltt,d)$\\
$\ell^0:=\ell(0,s_0)$,
$\ell^1:=\ell(1,s_0)$\\
If $d>1$ Then\\
\ii For $a\in\{0,1\}$\\
\ii \ii For $s\in\{0,1\}$\\
\ii \ii \ii
$\ell^{(s,a)}:=A(s,[x\ltt s_0],[y\ltt a],d\!-\!1)$\\
\ii \ii \ii $\ell^a:=\ell^a+\xi(s|s_0,a,x\ltt,y\ltt)\cdot
\ell^{(s,a)}$\\
Return $\min\{\ell^0,\ell^1\}$
\end{minipage}}
\caption{The \AIXI\ recursion for known loss matrix $\ell$.}
\label{fig:AIXI}
\end{center}
\end{minipage}

\vspace{1cm}
\begin{minipage}{0.7\textwidth}
\begin{center}
\fbox{
\begin{minipage}{\textwidth}
For $\tau=1,2,3,\ldots$\\
Sample $r_\tau\in\{0,1\}$ independ. s.t.
$P[r_\tau=1]=\gamma_\tau$\\
If $r_\tau=0$ Then\\
\ii Invoke subroutine $\FPL(\tau)$:\\
\ii \ii Sample $q_\tau^i\stackrel{d.}{\sim}\mbox{\textit{Exp}}$
independently for $1\leq i\leq n$\\
\ii \ii Select $I_\tau\fpl=\arg\min\limits_{1\leq i\leq
n}\{\eta_\tau\hat \Loss^i_{<\tau}+\ln w^i-q_\tau^i\}$\\
\ii$\langle$end of subroutine $\FPL(\tau)\rangle$\\
\ii Play $I_\tau\foe:=I_\tau\fpl$ for $B_\tau$ elementary time steps\\
\ii Set $\hat\loss_\tau^i=0$ for all $1\leq i\leq n$\\
Else\\
\ii Sample $I_\tau\foe\!\in\!\{1...n\}$ uniformly\\
\ii Play $I:=I_\tau\foe$
for $B_\tau$ elementary time steps\\
\ii Let $t_0(\tau):=\sum_{\tau'=1}^{\tau-1} B_{\tau'}$ and
$\loss_\tau^I:=\sum_{t=t_0(\tau)+1}^{t_0(\tau)+B_\tau}\loss_t^I$\\
\ii Let $\hat\loss_\tau^I=\loss_\tau^I n/\gamma_t$ and $\hat\loss_\tau^i=0$ for all $i\neq I$
\end{minipage}}
\caption{The algorithm $\FOE$. The parameters $\eta_t$, $\gamma_t$, and
$B_t$ will be specified in Theorem \ref{th:foe}.}
\label{fig:foe}
\end{center}
\end{minipage}
\end{center}
\end{figure}

\paranodot{Matrix games}
(as defined in Section \ref{secSim})
are straightforward in our setup. We just have to
consider that the opponent, i.e.\ the environment, does not
know our action $y_t$ when deciding its reaction
$x_t$: $\mu_i(x_t|x\ltt,y\leqt)=\mu_i(x_t|x\ltt,y\ltt)$.
\AIXI\ for $2\times 2$ matrix games
can then be implemented recursively as shown in Figure
\ref{fig:AIXI}, if we additionally assume that the
environments are \emph{Markov players} with two internal
states, corresponding to the reaction $x_t$ they are playing.
Since in step $t$, we don't know $x_t$ yet, \AIXI\ must
evaluate both $\mbox{AIXIrec}(0,x\ltt,y\ltt,d)$ and
$\mbox{AIXIrec}(1,x\ltt,y\ltt,d)$ and compute a weighted
mixture for both possible actions $a=0,1$. As long as we do
not yet know the loss matrix $\ell:(y_t,x_t)\mapsto[0,1]$
completely (which we assume to be deterministic), we
additionally compute an expectation over all assignments of
losses which are consistent with the history. To this aim, we
pre-define a finite set $\calL\subset\NNN$ which contains all
possible losses. In the simulations below, we use
$\calL=\{0\ldots 4\}\cup\{-16\}$, where the actual losses are
always in $\{0\ldots 4\}$ and the large negative value of
$-16$ encourages the agent to explore as long as he doesn't
know the losses completely. This is for obtaining interesting
results with moderate tree depth: otherwise, when the loss
observed by \AIXI\ is relatively low, \AIXI\ would explore
only with a large depth. This phenomenon is explained in
detail in Section \ref{secSim}.

%------------------------------%
\paranodot{Markov Decision Processes}
%------------------------------%
({\sc mdp}) have probably been most intensively studied from
all possible environments. In an {\sc mdp}, the environmental
behavior depends only on the last action and observation,
precisely $\mu(x_t|x\ltt,y\ltt)=\mu(x_t|x_{t-1},y_{t-1})$ in
case of a matrix game. For a $2\times 2$ game, a Markov player
is modelled by a $2\times 2\times 2$ transition matrix. It
turns out that the (uncountable) class of all transition
matrices with a uniform prior admits a closed-form solution:
\beqn
  \xi(x_t|x\ltt,y\ltt)={N_{x_{t\!-\!1}\to x_t}^{y_{t\!-\!1}}\!+\!1
  \over N_{x_{t\!-\!1}\to 0}^{y_{t\!-\!1}}\!+\!N_{x_{t\!-\!1}\to
1}^{y_{t\!-\!1}}\!+\!2},
\eeqn
where $N_{x_{t\!-\!1}\to x_t}^{y_{t\!-\!1}}$ counts how often in
the history the state $x_{t-1}$ transformed to state $x_t$
under action $y_{t-1}$. This is just Laplace' law of
succession \cite[Prob.2.11\&5.14]{Hutter:04uaibook}. (Observe
that $\xi$ is not Markov but depends on the full history.)
Note that the $\xi$ posterior estimate changes along the
expectimin tree (\ref{ydotxi}). Disregarding this important
fact as is done in Temporal Difference learning and variants
would result in greedy policies who have to rescue exploration
by ad-hoc methods (like $\eps$-greedy). One can show that
there exist self-optimizing policies $\tilde p$ for the class
of ergodic {\sc mdp}s \cite{Hutter:04uaibook}. Although the
class of transition matrices contains non-ergodic
environments, a variant of Theorem \ref{thSelfOpt} applies,
and hence the Bayes optimal policy $p^\xi$ is
self-optimizing for ergodic Markov players (which we will
exclusively meet). The intuitive reason is that the class is
compact and the non-ergodic environments have measure zero.

%%%%%%%%%%%%%%%%%%%%%%%%%%%%%%%%%%%%%%%%%%%%%%%%%%%%%%%%%%%%%%%
\section{Acting with Expert Advice (FoE)}\label{secFoE}
%%%%%%%%%%%%%%%%%%%%%%%%%%%%%%%%%%%%%%%%%%%%%%%%%%%%%%%%%%%%%%%

Instead of predicting or acting optimally with respect to a
model class, we may construct an agent from a class of
\emph{base agents}. We show how this can be accomplished for
fully active problems. The resulting algorithm will radically
differ from the \AIXI\ agent.

%------------------------------%
\paranodot{\emph{Prediction} with expert advice}
%------------------------------%
has been very popular in the last two decades. The base
predictors are called experts. Our goal is to design a master
algorithm which in each time step $t$ selects one expert $i$
and follows its advice (i.e.\ predicts as the expert does).
Thereby, we want to keep the master's \emph{regret}
$\ell\leqT^{\mbox{\scriptsize\textit{master}}}-\ell\leqT^*$
small, where $\ell\leqT^*$ is the cumulative loss of the best
expert in hindsight at time $T$. Usually, $T\geq 1$ not known
in advance. The state-of-the-art experts algorithms achieve
this: Loss bounds similar to (\ref{eq:lb}) can be proven, with
$\ell\leqT^\mu$ replaced by $\ell\leqT^*$ and $w^\mu$ replaced
by the prior weight of the best expert in hindsight, $w^*$.
These bounds hold \emph{in the worst case}, i.e.\ without any
assumption on the data generating process. In particular, the
environment which provides the feedback may be an adaptive
adversary. Since these bounds imply bounds in expectation in
the Bayesian setting (with slightly larger constants than
(\ref{eq:lb})), expert advice is in a sense the stronger
prediction strategy.

In order to protect against adaptive adversaries, we need to
randomize. In this work, we build on the \emph{Follow the
Perturbed Leader} $\FPL$ algorithm introduced by
\cite{Hannan:57}. (For space constraints, we won't discuss the
more popular alternative of weighted sampling at all.) We
don't even need to be told the true outcome after the master's
decision. All we need for the analysis is learning the
\emph{losses} of all experts, which are bounded wlog.\ in
$[0,1]$ (this is an important restriction which applies to all standard
experts algorithms). In this way, the master's actual decision
is based on the \emph{past cumulative loss} of the experts. A
key concept is that we must prevent the master from learning
too fast (or to slowly). This is achieved by introducing a
\emph{learning rate} $\eta_t$, which decreases to zero at an appropriate
rate with growing $t$.
%------------------------------%
%\paradot{Infinite \& universal expert classes}
%------------------------------%
Most of the literature assumes experts classes of finite size
$n$ with uniform prior $\frac{1}{n}$, in particular when the learning rate
$\eta_t$ is non-stationary. For $\FPL$, the case of arbitrary
non-uniform prior and countable expert
classes has been treated in \cite{Hutter:04expert}.%
\footnote{Given the large amount of recent literature, it should
be not too difficult to obtain similar assertions for WS
algorithms. However, as far as we know, the only result proven
up to now requires rapidly decaying weights \cite{Gentile:03},
which is therefore not appropriate for universal expert
classes.}

%------------------------------%
\paradot{Active problems}
%------------------------------%
In the passive \emph{full observation game} discussed so far
(i.e.\ we learn all losses), the notion of regret is not
problematic even against an adaptive adversary.\footnote{One
can even prove the following strong statement
\cite{Hutter:05expertx,Poland:05fpla}: If a strategy performs well
against an \emph{oblivious} adversary which does not at all
depend on our actions, then it also performs well against an
adaptive adversary.} However, the situation changes if the
reaction of the environment depends on our past actions.
Consider the simple case of two experts, one always suggesting
action 0 and the other one action 1. The environment is
reactive and ``unfair": Each expert incurs no loss as long as
we stay with its initial action (e.g. the action sequences
00000 and 111 have no loss). But as soon as we perform a
different action (e.g. 001), in all subsequent rounds both
experts incur loss 1. Each sensible strategy will soon explore
both actions, and compared to the pure experts, we incur large
loss. Consequently, we need to consider a different notion of
regret: Our performance is compared to what an expert could
achieve when he is actually put in our situation. In this
example, after the action sequence 001, we perform badly, but
so do all experts.

%------------------------------%
%\paradot{Reactive environments \& exploration}
%------------------------------%
Another problem with reactive environments is that we do not
necessarily get valid feedback for all experts in each round.
In the previous example, if we chose 0 as the first action and
learned that expert 0 had no loss at time $t=2$, it is not
legitimate to make any assumption on the loss of expert 1 at
$t=2$. Even if the environment tells us that the
\emph{pure} expert 1 had no loss, we are interested in
the loss of expert 1 put in our situation, i.e.\ after the
first action 0. But this loss we do not know. Precisely, we
know only the loss of an expert with the \emph{correct action
history} after the last time step in the past, where we (maybe
coincidentally) acted as he suggested. Instead of trying to
track the action history (which is possibly expensive), we
therefore use \emph{only the feedback from the currently
selected expert} $i$ and discard all other information. This
is commonly referred to as \emph{bandit setup}. Fortunately,
this issue can be successfully addressed by forcing
exploration, i.e.\ sampling according to the prior, with a
certain probability $\gamma_t$ \cite{McMahan:04}. This
\emph{exploration rate} $\gamma_t$ is decreased to zero
appropriately with growing $t$. Thus, in each time step we
decide to either \emph{follow} the perturbed leader or
\emph{explore}. Accordingly, we call our algorithm $\FOE$
(Follow or Explore). Bounds for the bandit setup are typically
similar to (\ref{eq:lb}), but with $-\ln w^*$ replaced by
(something larger than) $1/w^*$. Hence they are exponentially
larger in $w^*$, and one can show that this is sharp in
general.

%------------------------------%
\paradot{Increasing horizon}
%------------------------------%
If the environment is reactive, it is not sufficient to
consider only the short-term performance of the selected
expert $i$. This was first recognized by \cite{Farias:03}, who
considered the repeated game of ``Prisoner's Dilemma" and the
``tit for tat" opponent as a motivating example (see Section
\ref{secSim} for details). In this case, a good long term
strategy is cooperating (because of the particular opponent).
However defecting is dominant, i.e.\ the instantaneous loss of
defecting is \emph{always} smaller than that of cooperating.
So in order to notice that an expert (for instance the always
cooperating one) performs well, we have to evaluate it at
least over two time steps. In general, if we evaluate a chosen
expert over an
\emph{increasing number of time steps}, we hope that we
perform well in arbitrary reactive environments. This means
that the master works at a
\emph{different time scale}
$\tau$: in its $\tau$th time step, it gives the control to the
selected expert for $B_\tau\geq 1$ time steps (in the original
time scale $t$). As a consequence, the instantaneous losses
which the master observes are no longer uniformly bounded in
$[0,1]$, but in $[0,B_\tau]$. Fortunately, it turns out that the
analysis remains valid if $B_\tau$ grows unboundedly but
slowly enough. Only the convergence rate of the average
master's loss to the optimum is affected: we will obtain a
final rate of $t^{-\sfrac{1}{10}}$. The resulting algorithm
$\FOE$ (for a finite expert class) is specified in Figure
\ref{fig:foe} together with its subroutine $\FPL$. We may have
instantaneous and cumulative losses in both time scales, this
is always clear from the notation (e.g.\ $\ell^i_t$ vs.
$\hat\loss_{<\tau}^i$). Not surprisingly, most of $\FOE$ works
in the master time scale.

%------------------------------%
%\paradot{Improved FoE}
%------------------------------%
Note that $\FOE$ makes use of its observation \emph{only if he
decided to explore}, i.e.\ if $r_\tau=1$. This seems an
unnecessary waste of information. This is motivated from the
analysis, since $\FOE$ needs an \emph{unbiased loss estimate}
$\hat\loss$ (with respect to $\FOE$'s randomization). We just
chose the simplest way to guarantee this. For the simulations,
we concentrate on the following
\emph{faster learning variant}: approximate the probability
$p_\tau^i$ of the selected expert $i$ (jointly for exploration and
exploitation) by a Monte-Carlo simulation. Then always learn a
(close to) unbiased estimate
$\hat\loss_\tau^i=\frac{\ell_\tau^i}{p_\tau^i}$. The analysis of
$\FOE$ works in the same way for this modification, however
not resulting in better bounds. On the other hand, we will see
that modified
$\FOE$ learns faster.

%------------------------------%
%\paradot{FoE for infinitely many experts}
%------------------------------%
In case of a non-uniform prior and possibly infinitely many
experts, the exploration must be according to the prior
weights. This causes another problem: $\FOE$'s loss estimates
$\hat\loss$ need to be bounded, which forbids exploring experts
with very small prior weights. Hence we define for each expert
$i$, an \emph{entering time} $\calT^i\geq 1$ (at the master time
scale). Then $\FOE$ (including its subroutine $\FPL$) is
modified such that it uses only \emph{active} experts from
$\{i:\tau\geq\calT^i\}$. This guarantees additionally that
we have only a finite active set in each step, and the
algorithm remains computationally feasible.

\begin{Theorem}[Performance of FoE] \label{th:foe}
Assume $\FOE$ acts in an online decision problem with bounded
instantaneous losses
$\loss_t^i\in[0,1]$. Let the exploration rate
be $\gamma_\tau=\tau^{-\sfrac{1}{4}}$ and the learning rate be
$\eta_\tau=\tau^{-\sfrac{3}{4}}$. In case of uniform prior, choose
$B_\tau=\lfloor \tau^{\sfrac{1}{8}}\rfloor$.
In case of arbitrary prior let
$B_\tau=\lfloor t^{\sfrac{1}{16}}\rfloor$ and
$\calT^i=\lceil(w^i)^{-16}\rceil$.
Then in case of uniform prior, for all
experts $i$ and all
$T\geq 1$ we
have
\bqan
  \tfrac{1}{T}\Expect \Loss\foe\leqT &\leq& \tfrac{1}{T}\Loss\leqT^i+O(
  n^2 T^{-\sfrac{1}{10}}), \und\\
  \tfrac{1}{T}\Loss\foe\leqT &\leq& \tfrac{1}{T}\Loss\leqT^i+
  O(n^2 T^{-\sfrac{1}{10}}) \mbox{ w.p. } 1-T^{-2}.
\eqan
Consequently,
$\limsup_{T\to\infty} \tfrac{1}{T}(\Loss\foe\leqT-\loss^i\leqT)\leq0$
a.s. For non-uniform prior, corresponding assertions hold with
$O$-terms replaced by
$O\big(T^{-\sfrac{1}{10}}+(w^i)^{-22}T^{-1}\big)$.
\end{Theorem}

The proof of this main theorem on the performance of
$\FOE$ can be found in \cite{Poland:05dule}. Similar bounds
hold for larger $B_\tau<\tau^{\frac{1}{4}}$. These bounds are improvable
\cite{Poland:05fpla}, and it is possible to prove any regret bound
$O\Big((\frac{1}{w^i})^{c}+(\log\frac{1}{w^i})T^{\frac{2}{3}+\eps}\Big)$, at
the cost of increasing $c$ where $\eps\to 0$.
In the simulations, we used $B_\tau=\tau^{0.24}$ for faster learning.
For playing $2\times 2$ matrix games, we will use the class of all
16 deterministic four-state Markov experts. That is, each
expert consists of a lookup table with all the actions for
each of the 4 possible combination of moves in the last round.
In the first round, the expert plays uniformly random.
(Compare the standard results on learning matrix games with
expert advice by \cite{Freund:99}.)

%%%%%%%%%%%%%%%%%%%%%%%%%%%%%%%%%%%%%%%%%%%%%%%%%%%%%%%%%%%%%%%
\section{Simulations}\label{secSim}\label{sec:PD}
%%%%%%%%%%%%%%%%%%%%%%%%%%%%%%%%%%%%%%%%%%%%%%%%%%%%%%%%%%%%%%%

As already indicated, it is our goal to explore and compare
the performance of the two universal learning approaches
presented so far, in particular for problems which are not
solved by passive or greedy learners. To this aim, repeated
$2\times 2$ matrix games are well suited:
\begin{itemize}
\item they are simple, such that (close to) universal learning
is computationally feasible even with brute-force
implementation;
\item they provide situations where optimal long-term behavior
significantly differs from greedy behavior (e.g.\ Prisoner's
Dilemma);
\item moreover, we can observe how universal learners can
exploit potentially weak adversaries;
\item and finally, we can test the two universal learners
against each other.
\end{itemize}
We begin by describing the experimental setup and the
universal learners. After that, we will discuss five
$2\times 2$ matrix games, presenting experimental results and
highlighting their interesting aspects.

%------------------------------%
\paradot{Setup}
%------------------------------%
A $2\times 2$ matrix game consists of two matrices
$R_1,R_2\in\RRR^{2\times 2}$, the first one containing rewards for the row
player, the second one rewards for the column player.
(It does not cause any problem that for convenience, we have
developed the theory in terms of \emph{losses} rather than
rewards: one may be transformed into the other by simply
inverting the sign. So for the discussion of the results, we
will keep the rewards, as this is more standard in game
theory.)
A \emph{single game} proceeds in the following way: the first
player chooses a row action $i\in\{0,1\}$ and simultaneously
the second a column action $j\in\{0,1\}$, both
players without knowing the opponent's move. Then reward
$R_k(i,j)$ is payed to player $k$ ($k=1,2$), and $i$ and $j$
are revealed to both players. A repeated game consists of $T$
single games. We chose $T=20000$, if at least one opponent is
$\FOE$ (which has slow learning dynamics, as we will see),
and $T=100$ for the fast learning \AIXI\ (unless it is plotted
in the same graph as $\FOE$). If at least one randomized
player participates, the run is repeated 10 times, and usually
the average is shown.
We will consider only \emph{symmetric} games, where one
player, when put in the position of the other player (i.e.\
when exchanging $R_1$ and $R_2$), has a symmetric strategy
(maybe after exchanging the actions). We will meet precisely
three types of symmetry: in the ``Matching Pennies", $R_1=R_2$
after inverting the action of the row player, and in the
``Battle of Sexes" game, $R_1=R_2$ after inverting both
players' actions, and in all other games we $R_1=R_2^T$
(transpose). In these latter games, we will call the action 0
``defect" and 1 ``cooperate".
All games we consider have rewards in $\{0\ldots 4\}$. The
\AIXI\ and $\FOE$ agents are used as specified in the previous
sections, with the classes of all two-state Markov
environments and all deterministic four-state Markov experts,
respectively. For \AIXI, we will concentrate the presentation
on the almost consistent horizon variant, since it performs
always better than the moving horizon variant. For $\FOE$, we
will concentrate on the faster learning variant.

\begin{figure}
\begin{center}
\begin{tabular}{| r | c | c |}
\hline
& \quad 0 \quad & 1\\
\hline
0 & 1,1 & 4,0\\
\hline
1 & 0,4 & 2,3\\
\hline
\multicolumn{3}{c}{Prisoner's Dilemma}
\end{tabular}
\quad
\begin{tabular}{| r | c | c |}
\hline
& 0 & 1\\
\hline
0 & 2,2 & 3,0\\
\hline
1 & 0,3 & 4,4\\
\hline
\multicolumn{3}{c}{Stag Hunt}
\end{tabular}
\quad
\begin{tabular}{| r | c | c |}
\hline
& 0 & 1\\
\hline
0 & 0,0 & 4,1\\
\hline
1 & 1,4 & 2,2\\
\hline
\multicolumn{3}{c}{Chicken}
\end{tabular}
\\[1ex]
\begin{tabular}{| r | c | c |}
\hline
& \quad 0 \quad & 1\\
\hline
0 & 2,4 & 0,0\\
\hline
1 & 0,0 & 4,2\\
\hline
\multicolumn{3}{c}{Battle of Sexes}
\end{tabular}
\quad
\begin{tabular}{| r | c | c |}
\hline
& \quad 0 \quad & 1\\
\hline
0 & 4,0 & 0,4 \\
\hline
1 & 0,4 & 4,0\\
\hline
\multicolumn{3}{c}{Matching Pennies}
\end{tabular}
\caption{Reward Matrices}
\end{center}
\end{figure}

%------------------------------%
\paradot{Prisoner's Dilemma}
%------------------------------%
This dilemma is classical. The reward matrices are $R_1={1
4\choose 0 3}$ and $R_2=R_1^T$, with the following
interpretation: The two players are accused of a crime they
have committed together. They are being interrogated
separately. Each player can either cooperate with the other
player (don't tell the cops anything), or he defects (tells
the cops everything but blame the colleague). The punishments
are according to the players' joint decision: if none of them
gives evidence, both get a minor sentence. If one gives
evidence and the other one keeps quiet, the traitor gets free,
while the other gets a huge sentence. If both give evidence,
then they both get a significant sentence. (There is also an
easier variant ``Deadlock" which we not discuss.)

It is clear that giving evidence, i.e.\ defecting, is an
instantaneously \emph{dominant} action: regardless of what the
opponent does, the immediate reward is always larger for
defecting. However, if both players would agree to cooperate,
this is the ``social optimum" and guarantees the better
long-term reward in the repeated game. A well-known instance
for this case is playing against the ``tit for tat" strategy
strategy which cooperates in the first move and subsequently
performs the action we did in the previous move. Similar but
harder to learn are ``two tit for tat" and ``three tit for
tat", which defect in the first move and cooperate only if we
cooperated two respectively three times in a row. Note that
although ``two tit for tat" and ``three tit for tat" are not
in \AIXI's model class, probabilistic versions of the
strategies are: if the probability of ``adversary defected, I
cooperate, then the adversary will cooperate in the next
round" is chosen correctly (namely $\frac{1}{2}$ for 2-tit for
tat and $\approx 0.57$ for 3-tit for tat), then the expected
number of rounds I have to cooperate until the adversary will
do so is 2 respectively 3.

\begin{figure}[t!]
\begin{minipage}{\graphwidth}
\begin{center}
\epsfig{file=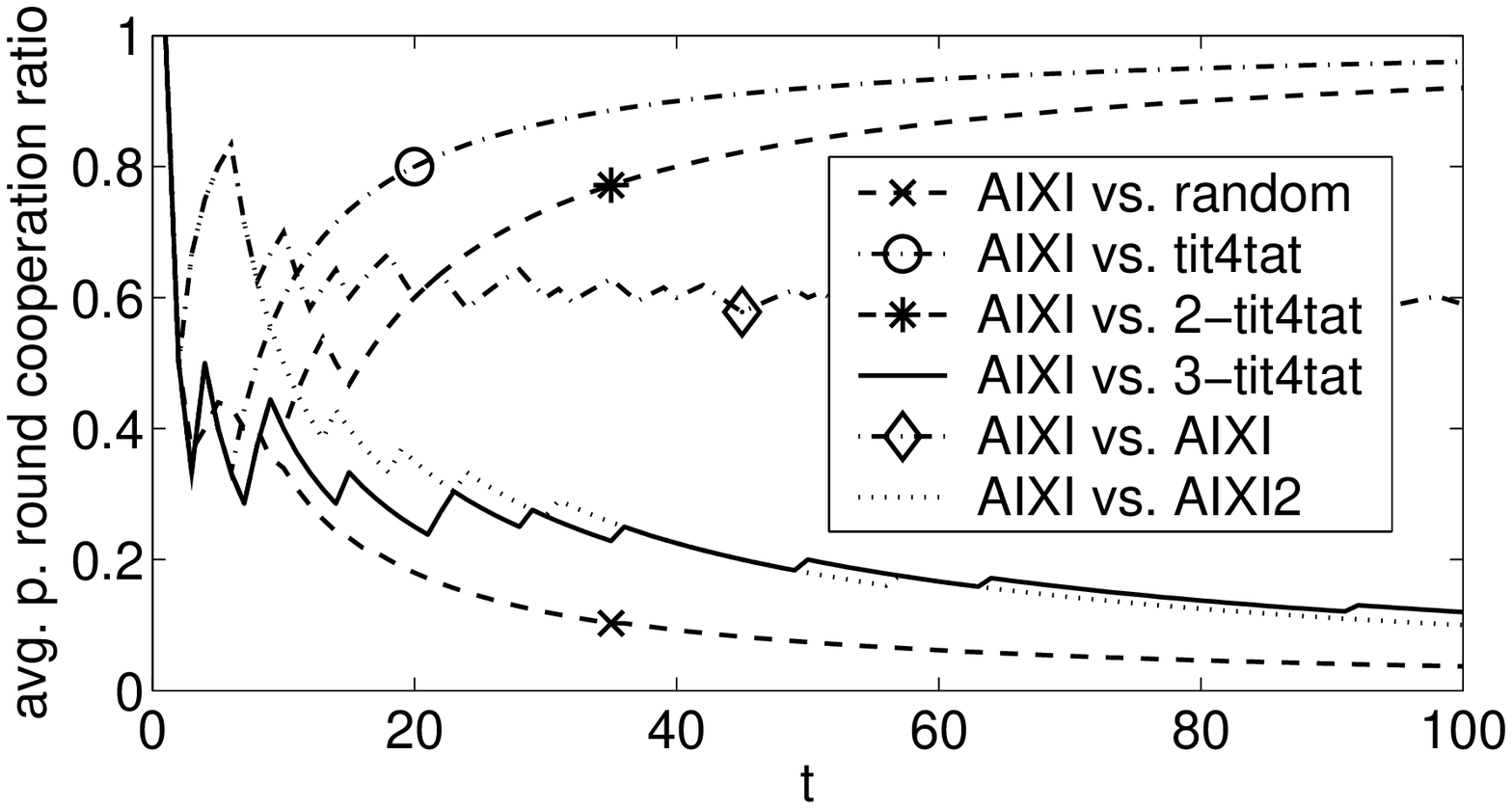, width=\graphwidth}\vspace*{-3ex}
\caption{\AIXI\ in Prisoner's Dilemma}
\label{fig:PDaixi}
\end{center}
\end{minipage}
\hfill%
\begin{minipage}{\graphwidth}
\begin{center}
\epsfig{file=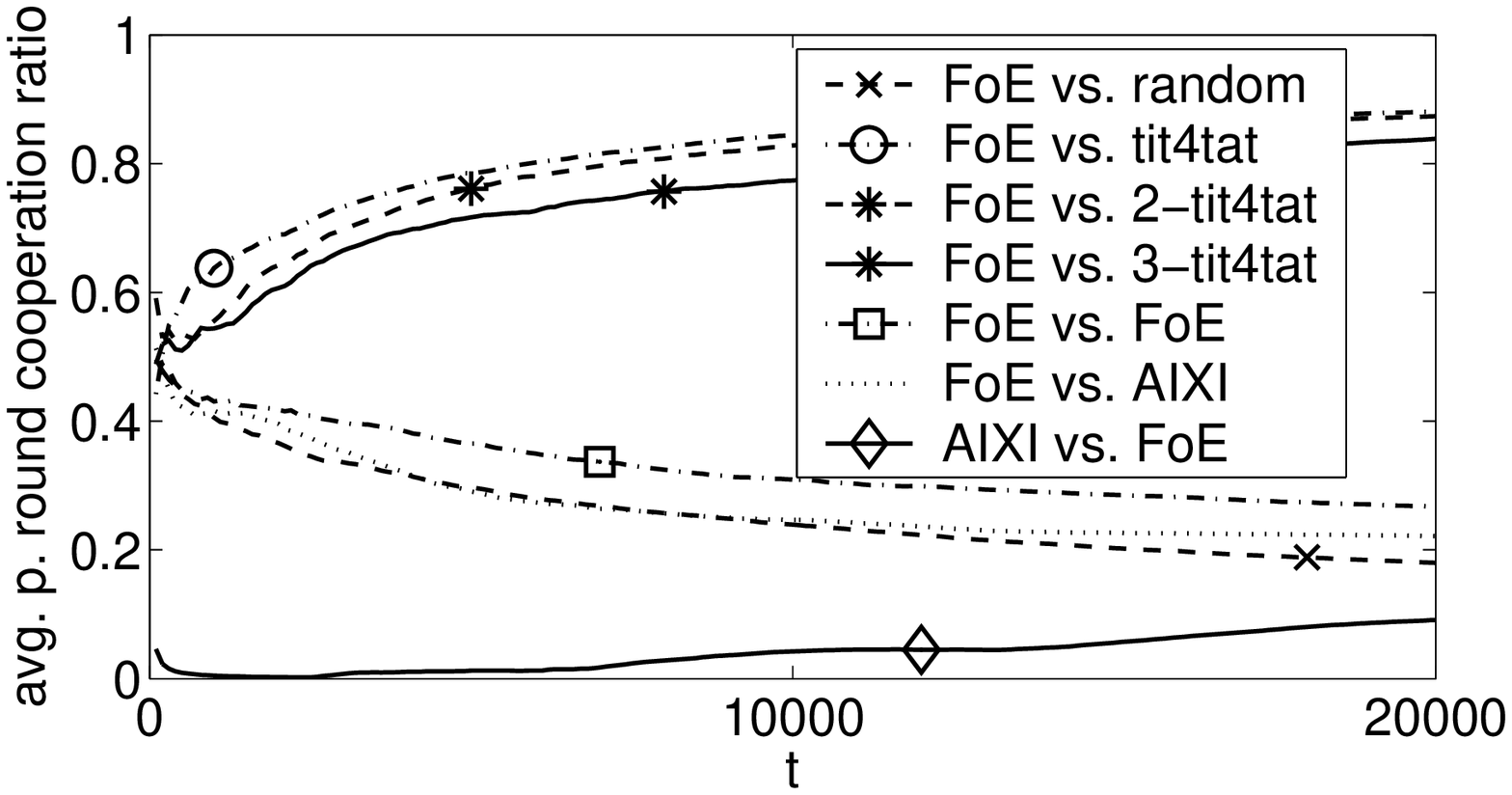, width=\graphwidth}\vspace*{-3ex}
\caption{$\FOE$ (and \AIXI) in Prisoner's Dilemma}
\label{fig:PDfoe}
\end{center}
\end{minipage}
\end{figure}

Figure \ref{fig:PDaixi} shows that in most cases, \AIXI\
learns very quickly the best actions. (This is the consistent
horizon variant, the moving horizon variant will be discussed
with the next game, Stag Hunt.) If the opponent is memoryless
as for example the uniform random player, \AIXI\ constantly
defects after short time. Against tit for tat and two tit for
tat, \AIXI\ cooperates after short time. The figure shows the
average per round rate of cooperation, which after a few
exploratory moves converges to the optimal action as
$\frac{1}{t}$. However, \AIXI\ does not learn to cooperate
against three tit for tat. The reason is the general problem
that in order to increase exploration, \AIXI\ needs
exponential depth of the expectimin tree. Assume that a
certain action sequence of length $n$ is favorable against the
true environment, which has however not too high a current
weight. In this instance, cooperating three times in a row is
favorable against (the probabilistic version of) 3-tit for
tat. In order to recognize that this is worth exploring,
\AIXI\ has to build a branch of depth $n=3$ in the expectimin
tree, which has (because of the relatively low prior weight)
very small probability $\sim\exp(-n)$ however. Then it needs
an exponentially large subtree below this branch to accumulate
enough (virtual) reward in order to encourage exploration.

One more problem arises when \AIXI\ plays against another \AIXI.
Here, the perfectly symmetric setting results in both playing the
same actions in each move, hence they are not correctly learning.
We might try to remedy this by varying the tree depth of the
second \AIXI\ (denoted \AIXI2 in the figure), however it turns out
that in this case, both \AIXI's do not learn at all to cooperate
(see \cite[Sec.8.5.2]{Hutter:04uaibook} for a possible reason).

We now turn to the performance of $\FOE$ (the faster learning
variant) as evaluated in Figure \ref{fig:PDfoe}. As expected,
$\FOE$ learns much slower than \AIXI\ (note the different time
scale). On the other hand, its exploration is strong enough to
learn 3-tit for tat (and even harder instances). When playing
against another instance of $\FOE$, we notice however that
they usually do not succeed to overcome the dominance of
mutual defection. Also when $\FOE$ competes with \AIXI, they
tend to learn mutual defection rather than cooperation.
(Sometimes, they learn cooperation in one or two of the
possible states of the MDP.)

%

%------------------------------%
\paradot{Stag Hunt}
%------------------------------%
This game is also known as ``Assurance". The reward matrices
are $R_1={2 3\choose 0 4}$ and $R_2=R_1^T$. Two players are
hunting together. If they cooperate, they will catch the stag.
However, one player might not trust the other, in which case
he chases rabbits on his own instead. In this case, the other
one won't get anything if he tries to cooperate. If both
defect, then they are in conflict, and each player gets less
rabbits. Although the optimum for both players is to
cooperate, they need to trust each other sufficiently. If one
player plays uniformly random, it is better for the other to
go for the rabbits. Also, defecting has the lower variance.

\begin{figure}[t!]
\begin{minipage}{\graphwidth}
\begin{center}
\epsfig{file=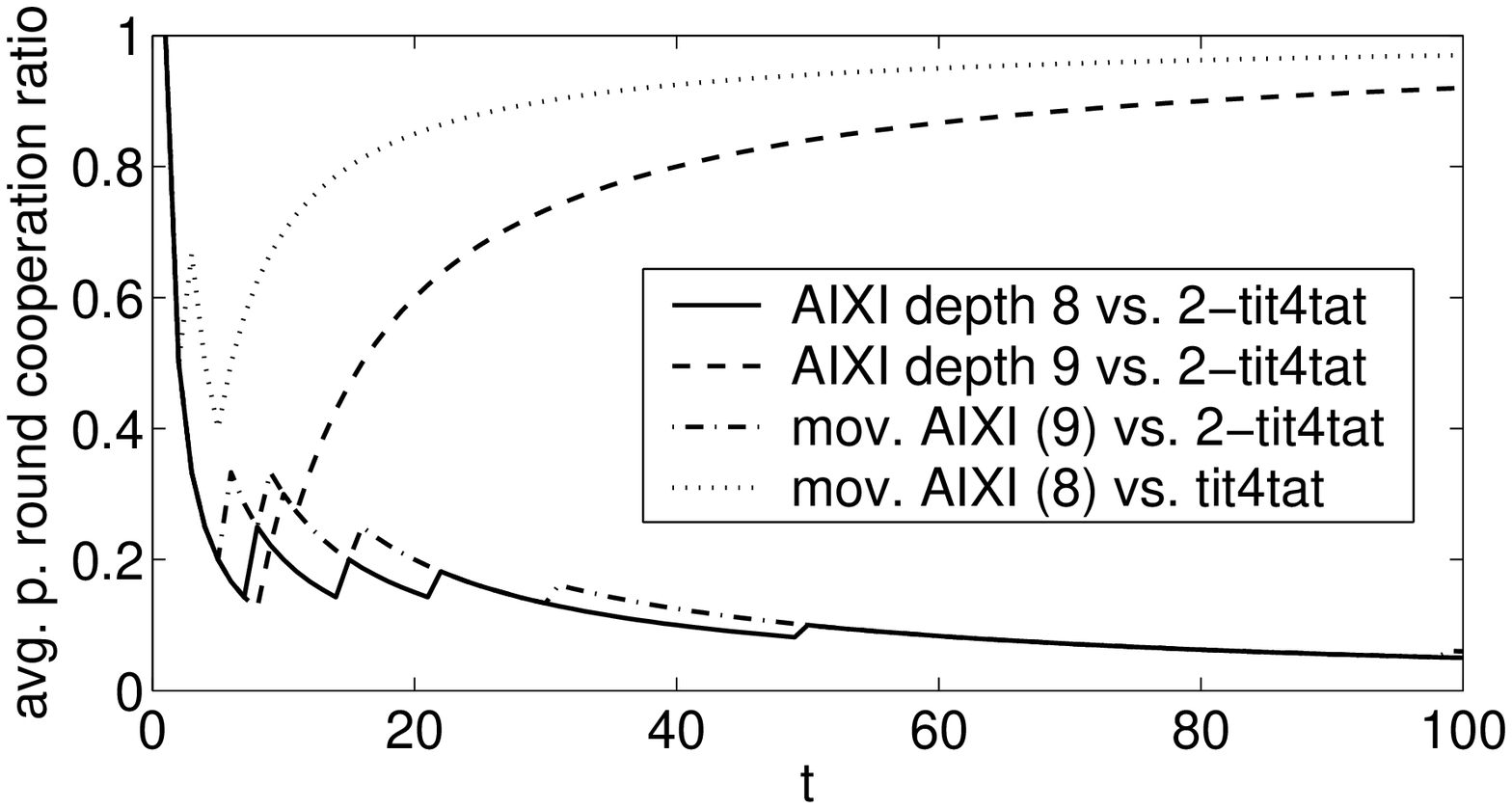, width=\graphwidth}\vspace*{-3ex}
\caption{Stag Hunt: \AIXI\ and its moving horizon variant}
\label{fig:SHaixi}
\end{center}
\end{minipage}
\hfill%
\begin{minipage}{\graphwidth}
\begin{center}
\epsfig{file=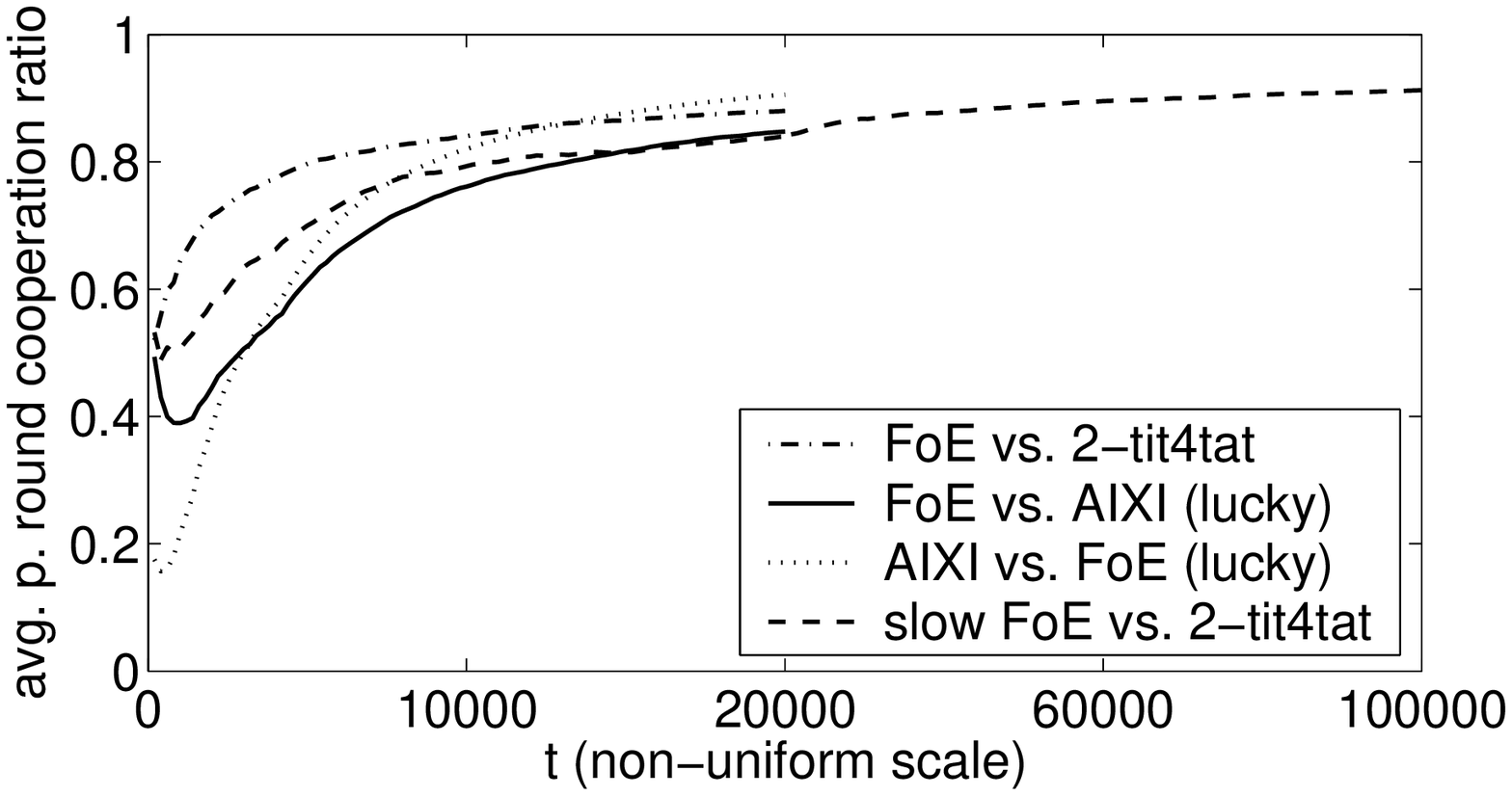, width=\graphwidth}\vspace*{-3ex}
\caption{Stag Hunt: $\FOE$ and its slower learning variant}
\label{fig:SHfoe}
\end{center}
\end{minipage}
\end{figure}

Maybe it is surprising to observe that \AIXI\ (with a depth of
8) does not learn to cooperate against 2-tit for tat (Figure
\ref{fig:SHaixi}). The reason is that defecting has a
relatively good payoff, and therefore exploration is not
encouraged as discussed in the previous subsection. If the
depth of the tree is increased to 9, \AIXI\ learns cooperation
against 2-tit for tat (but not against 3-tit for tat). We also
see that the moving horizon variant of \AIXI\ has even more
problems with exploration: It does not learn cooperating
against 2-tit for tat, even with depth 9. The explanation is
that even if \AIXI\ decides to explore in one time step, in
the next step this exploration might not be correctly
continued, as the tree is now explored to a different level.
This observation can also be made for the Prisoner`s Dilemma.
In fact, the consistent horizon variant performs \emph{always}
better than moving horizon.

As before, $\FOE$ learns much slower but explores more
robustly (Figure \ref{fig:SHfoe}), neither 2- nor 3-tit for
tat are a problem. Unlike in the Prisoner's Dilemma, if \AIXI\
and $\FOE$ are competing, they learn mutual cooperation in
almost half of the cases, an average over such lucky instances
is given in the figure. The same is valid for $\FOE$ against
$\FOE$, while \AIXI\ against \AIXI\ has the same symmetry problem
as already observed in the Prisoner's Dilemma. The original
slower learning variant of $\FOE$ reaches the same average
level of performance only after $10^5$ time steps instead of
$2\cdot 10^4$ steps, and moreover with a variance twice as
high.

%------------------------------%
\paradot{Chicken}
%------------------------------%
The reward matrices $R_1={0 4\choose 1 2}$ and $R_2=R_1^T$ of
the ``Chicken" game (also known as ``Hawk and Dove") can be
interpreted as follows: Two coauthors write a paper, but each
tries to spend as little effort as possible. If one succeeds
to let the other do the whole work, he has a high reward. On
the other hand, if no one does anything, there will be no
paper and thus no reward. Finally, if both decide to
cooperate, both get some reward. Here, in the repeated game,
it is socially optimal to take turns cooperating and
defecting.\footnote{We assume that the authors are not very
good at cooperating, and that the costs of cooperating more
than compensate for the synergy. We could assign a reward of 3
instead of 2 to mutual cooperation. This is the less
interesting situation of ``Easy Chicken", where cooperating is
the optimal long-term strategy like in the previous games.}
Still the best situation for one player is if he emerges as
the ``dominant defector", defecting in most or all of the
games, while the other one cooperates.

\begin{figure}[t!]
\begin{minipage}{\graphwidth}
\begin{center}
\epsfig{file=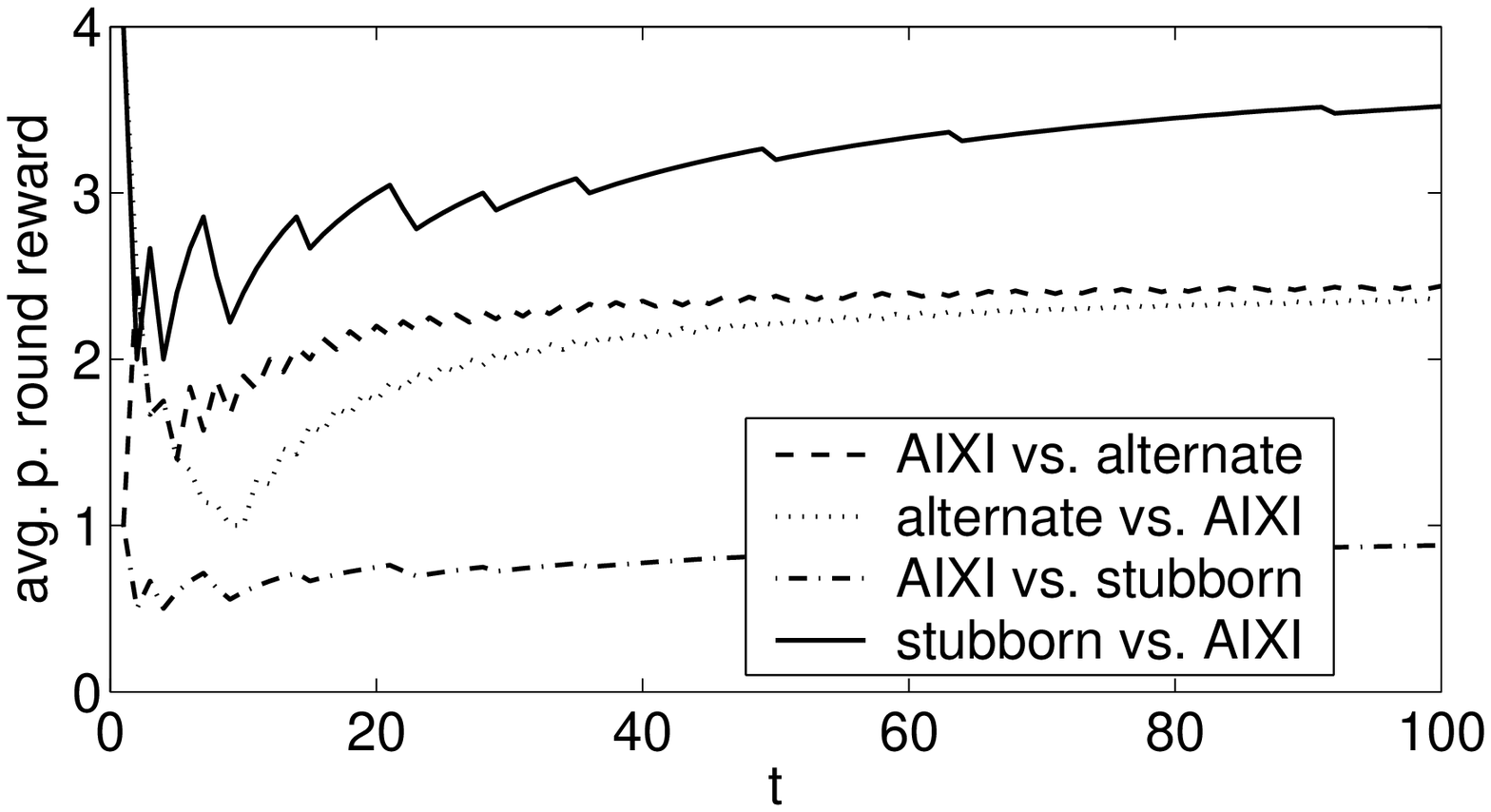, width=\graphwidth}\vspace*{-3ex}
\caption{\AIXI\ in the Chicken game}
\label{fig:ChickenAixi}
\end{center}
\end{minipage}
\hfill%
\begin{minipage}{\graphwidth}
\begin{center}
\epsfig{file=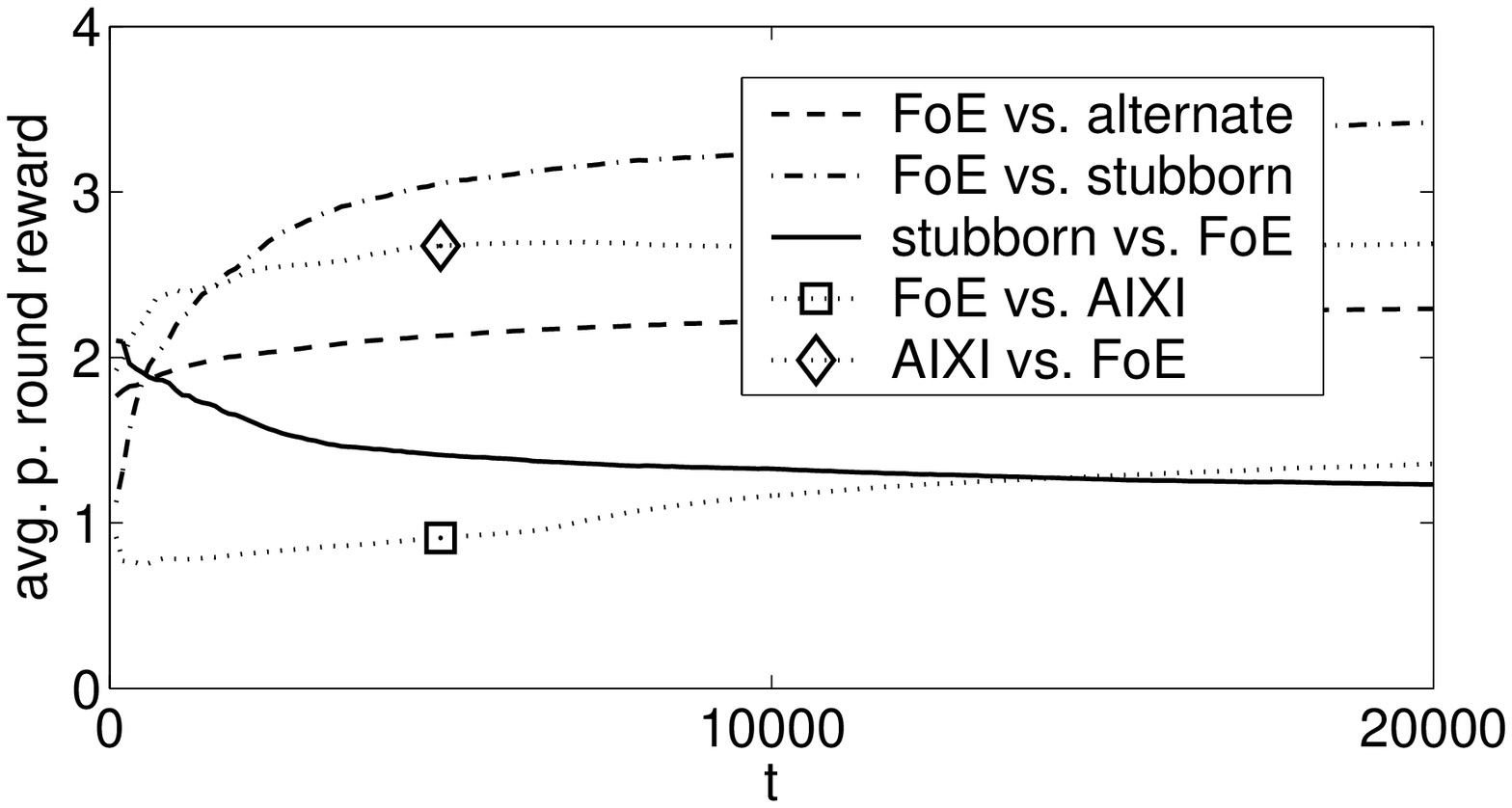, width=\graphwidth}\vspace*{-3ex}
\caption{$\FOE$ (and \AIXI) in the Chicken game}
\label{fig:ChickenFoe}
\end{center}
\end{minipage}
\end{figure}

If the opponent steadily alternates between cooperating and
defecting, then \AIXI\ quickly learns to adapt. This can be
observed in Figure \ref{fig:ChickenAixi}, where the
performance is given in terms of average per round reward
instead of cooperation rate. However, \AIXI\ is not obstinate
enough to perform well against a ``stubborn" adversary that
would cooperate only after his opponent has defected for three
successive time steps. Here, \AIXI\ learns to cooperate,
leaving his opponent the favorable role as the dominant
defector. (However, \AIXI\ learns to dominate the less
stubborn adversary which cooperates after two defecting
actions.) When two \AIXI s play against each other, they again
have the symmetry problem. Interestingly, if we break symmetry
by giving the second \AIXI\ a depth of 9, he will turn out the
dominant defector (not shown in the graph).

$\FOE$ behaves differently in this game (Figure
\ref{fig:ChickenFoe}). While he learns to deal with the
steadily alternating adversary and emerges as the dominant
defector against the stubborn one, he would give precedence to
\AIXI\ in most cases. This is not hard to explain, since $\FOE$
in the beginning plays essentially random. Thus \AIXI\ learns
quickly to defect, and for $\FOE$ remains nothing but learning
to cooperate. However, this does not always happen: In the
minority of the cases, $\FOE$ defects enough such that \AIXI\
decides to cooperate, and $\FOE$ will be the dominant
defector. (Hence the average shown in the graph is less clear
in favor of \AIXI.)

%------------------------------%
\paradot{Battle of Sexes}
%------------------------------%
In this game, a married couple wants to spend the evening
together, but they didn't settle if they would go to the
theater (her preference) or the pub (his preference). However,
if they fail to meet, both have a boring evening (and no
reward at all). The reward matrices are $R_1={2 0\choose 0 4}$
and $R_2={4 0\choose 0 2}$. Coordination is clearly important
in the repeated game. Like in ``Chicken", taking turns is a
social optimum, while it is best for one player if his choice
becomes dominant.

\begin{figure}[t!]
\begin{minipage}{\graphwidth}
\begin{center}
\epsfig{file=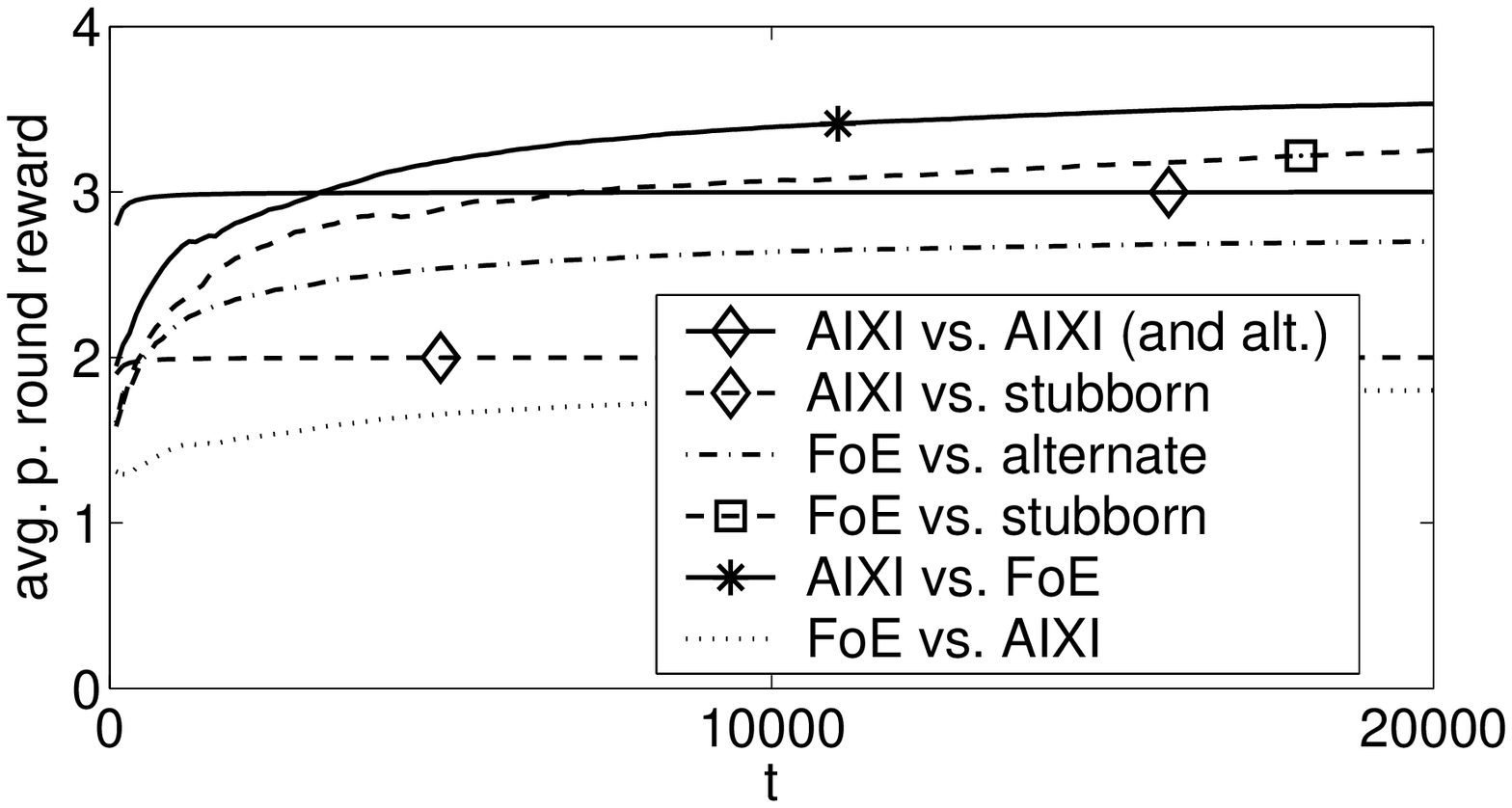, width=\graphwidth}\vspace*{-3ex}
\caption{\AIXI\ and $\FOE$ in Battle of Sexes}
\label{fig:BoS}
\end{center}
\end{minipage}
\hfill
\begin{minipage}{\graphwidth}
\begin{center}
\epsfig{file=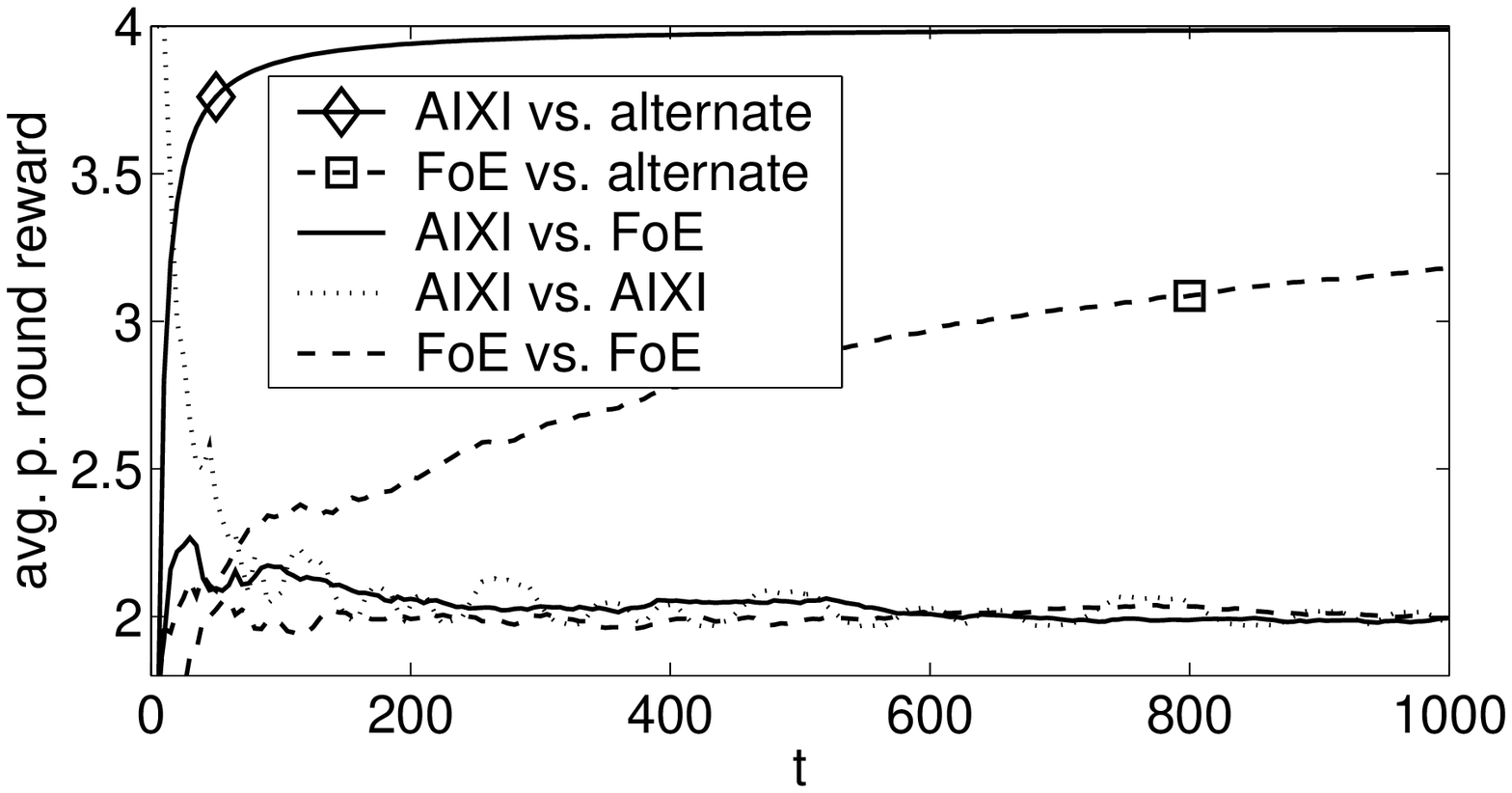, width=\graphwidth}\vspace*{-3ex}
\caption{\AIXI\ and $\FOE$ in Matching Pennies}\vspace{-3ex}
\label{fig:MP}
\end{center}
\end{minipage}
\end{figure}

In Figure \ref{fig:BoS}, our universal learners show similar
performance like in the Chicken game. Both learn to deal with
an alternating partner. $\FOE$ also learns to dominate over a
stubborn adversary which plays his less favorite action only
after the opponent insists three times on that. \AIXI\ is
dominated by this stubborn player. However, \AIXI\ always
dominates $\FOE$. Finally, in contrast to the Chicken game,
\AIXI\ against \AIXI\ does not have the symmetry problem, but they
both learn to alternate.

%------------------------------%
\paradot{Matching Pennies}
%------------------------------%
Each player conceals in his palm a coin with either heads or
tails up. They are revealed simultaneously. If they match
(both heads or both tails), the first player wins, otherwise
the second. This is the only zero-sum game of the games we
consider, where $R_1={4 0\choose 0 4}$ and $R_2={0 4\choose 4
0}$. Thus, there is a minimax strategy for both players, which
is actually uniform random play. On the other hand,
deterministic repeated play is potentially exploitable by the
adversary.

Figure \ref{fig:MP} shows the results for this last game we
present. Both \AIXI\ and $\FOE$ learn to exploit a predictable
adversary, namely the player alternating between 0 and 1. The
other games are balanced in the long run, only in the
beginning \AIXI\ succeeds to exploit $\FOE$ a little. If two
\AIXI s compete, it is important to break symmetry, then
both learn to alternate (this situation is shown in the
graph). If symmetry is not broken, the row player (who tries
to match) always wins.

%%%%%%%%%%%%%%%%%%%%%%%%%%%%%%%%%%%%%%%%%%%%%%%%%%%%%%%%%%%%%%%
\section{Discussion}\label{secDC}
%%%%%%%%%%%%%%%%%%%%%%%%%%%%%%%%%%%%%%%%%%%%%%%%%%%%%%%%%%%%%%%

Altogether, universal learners perform well in repeated matrix
games. They usually learn to prefer the optimal long-term
action to greedy behavior (Prisoner's Dilemma and Stag Hunt).
If possible they are able to exploit a predictable adversary
(Matching Pennies). And they learn good strategies when it is
necessary to foresee the opponent's action (Chicken and Battle
of Sexes). Of the two approaches we presented and compared,
\AIXI\ learns much faster than $\FOE$, but $\FOE$ explores more
thoroughly. Of course, there is a trade-off between
exploration and fast learning. Interestingly, it may depend on
the adversary (and thus on the environment) if fast learning
or exploration is the better \emph{long-term} strategy: In
Chicken and Battle of Sexes, \AIXI\ profits against $\FOE$ by
learning fast and dictating its preferred action, but looses
against the stubborn opponent because of not exploring enough.

%------------------------------%
\paradot{Acknowledgement}
%------------------------------%
This work was supported by the Swiss NSF grant 2100-67712.

%%%%%%%%%%%%%%%%%%%%%%%%%%%%%%%%%%%%%%%%%%%%%%%%%%%%%%%%%%%%%%%
%         Bibliography        %
%%%%%%%%%%%%%%%%%%%%%%%%%%%%%%%%%%%%%%%%%%%%%%%%%%%%%%%%%%%%%%%

\bibliographystyle{alpha}

\end{document}